\documentclass[conference]{IEEEtran}
\IEEEoverridecommandlockouts
\usepackage{cite}
\usepackage{amsmath,amssymb,amsfonts}
\usepackage{algorithmic}
\usepackage{graphicx}
\usepackage{textcomp}
\usepackage{xcolor}
\usepackage{float}
\usepackage{multirow}
\usepackage{paralist}
\usepackage{tabularx}
\usepackage{comment}
\def\BibTeX{{\rm B\kern-.05em{\sc i\kern-.025em b}\kern-.08em
    T\kern-.1667em\lower.7ex\hbox{E}\kern-.125emX}}
\begin{document}

\title{Leveraging Transformers for StarCraft Macromanagement Prediction}

\author{\IEEEauthorblockN{Muhammad Junaid Khan}
\IEEEauthorblockA{\textit{Department of Computer Science} \\
\textit{University of Central Florida}\\
Orlando, USA \\
junaid\_k@knights.ucf.edu}
\and
\IEEEauthorblockN{Shah Hassan}
\IEEEauthorblockA{\textit{Department of Computer Science} \\
\textit{University of Central Florida}\\
Orlando, USA \\
shahhassan@knights.ucf.edu}
\and
\IEEEauthorblockN{Gita Sukthankar}
\IEEEauthorblockA{\textit{Department of Computer Science} \\
\textit{University of Central Florida}\\
Orlando, USA \\
gitars@eecs.ucf.edu}
}

\maketitle

\begin{abstract}
Inspired by the recent success of transformers in natural language processing and computer vision applications, we introduce a transformer-based neural architecture for two key StarCraft II (SC2) macromanagement tasks: global state and build order prediction.
Unlike recurrent neural networks which suffer from a recency bias, transformers are able to capture patterns across very long time horizons, making them well suited for full game analysis. Our model utilizes the MSC (Macromanagement in StarCraft II) dataset and improves on the top performing gated recurrent unit (GRU) architecture in predicting global state and build order as measured by mean accuracy over multiple time horizons. We present ablation studies on our proposed architecture that support our design decisions.

One key advantage of transformers is their ability to generalize well, and we demonstrate that our model achieves an even better accuracy when used in a transfer learning setting in which models trained on games with one racial matchup (e.g., Terran vs. Protoss) are transferred to a different one.  
We believe that transformers' ability to model long games, potential for parallelization, and generalization performance make them an excellent choice for StarCraft agents.
\end{abstract}


\section{Introduction}
Real-time strategy (RTS) games such as StarCraft II pose multiple interesting challenges for AI-based agents.  Rather than tackling full game play, many systems focus on the \textbf{micromanagement} aspects of the game such as moving, targeting, and resource gathering; these tactical decisions only require reasoning over a short-time horizon and are more amenable to machine learning approaches.  In contrast, \textbf{macromanagement} addresses  strategic gameplay choices such as production decisions and countering opponents' expansion. Ontañón et al.~\cite{hc_feat1} note that “a good macro player has the larger army” whereas “a good micro player keeps their units alive for a longer amount of time”. 

The MSC (Macromanagement in StarCraft II)  dataset~\cite{msc} was created specifically to address the challenge of developing machine learning agents with human-level macromanagement performance.  They identify two key prediction tasks that are important for tracking the high-level flow of the game:
\begin{compactitem}
\item \textbf{global state prediction}: predicting whether the current game state will lead to a win or a loss for the player. This is important for evaluating the comparative benefits of different strategic choices. 
\item \textbf{build order prediction}: predicting which combat units will be researched, produced, and updated. Early prediction of the opponent's future army composition is a competitive advantage when making production decisions. 
\end{compactitem}
Vinyals et al.\cite{chal} note that win-loss prediction in SC2 is challenging even for expert human players. The fog-of-war in SC2 that blocks players from seeing the map until they explore or build an area makes the current state only partially observable, rendering these prediction problems more difficult, as does the large state-action space~\cite{risi,Xu_2019}.

This paper introduces a transformer-based approach to global state and build order prediction that is inspired by the recent success of transformers in various natural language processing \cite{attn} and computer vision  tasks  \cite{detr,vitt}.  Unlike most sequence to sequence learning models, the transformer does not use a recurrent structure to capture time dependencies in sequences.  Instead it uses an attention mechanism to decide which parts of the sequence are important~\cite{attn}. A positional encoding is added to the state embedding to preserve the ordering information. 

Recurrent neural networks suffer from a recency bias, treating recent information as more important than actions in the distant past~\cite{recencybias}.  Even gated units have difficulty retaining information over a long time horizon, since they have a probability of forgetting information at every time step.  This property is useful for many tasks but poses a problem for macromanagement, which relies on the cumulative effects of decisions over a very long time horizon.   
We hypothesize the following:
\begin{compactitem}
\item \textbf{H1}: A transformer will outperform an architecture that depends on gated recurrent units (GRUs) to learn temporal dependencies.
\item \textbf{H2}: A simple self-attention architecture will achieve equivalent performance to the GRU.
\item \textbf{H3}: Due to its generalization power~\cite{generalization}, a transformer will be highly effective at transfer learning tasks such as learning from observing the outcome of one racial matchup (e.g. Protoss vs. Zerg) how to predict the outcome of different matchups.
\end{compactitem}

This paper also presents an ablative study evaluating the benefits of different design decisions, such as the usage of skip connections and the inclusion of decoder layers.  We show that the best version of our transformer outperforms a competitive benchmark using GRUs (\textbf{H1}).  The next section describes related work on the applications of deep learning architectures to SC2. 

\section{Related Work}
Several early systems explored the application of player modeling techniques in StarCraft towards the aim of improving strategic play.
BroodWarBotQ was an early Bayesian system that explicitly integrated opening and technology tree prediction into a competition bot~\cite{synnaeve2011}.  However most of the current deep learning systems utilize reinforcement learning and do not explicitly separate prediction from decision-making. Reinforcement learning techniques often treat win/loss as the primary reward signal, and learn value functions that implicitly predict game outcome. 

\subsection{Deep Reinforcement Learning}
Vinyals et al.\cite{chal} stimulated much work in this area by introducing SC2LE, a SC2 environment for reinforcement learning, They note the relative ease of creating RL agents from scratch to learn the mini games, which test micromanagement skills, whereas they advocate the use of  replay datasets as a better way to achieve proficiency on the full game. Tang et al. used this environment to demonstrate an RL agent that could do macromangement using Convolutional Neural Network Fitted Q-Learning (CNNFQ) \cite{tang}.

Xu et al. \cite{Xu_2019} created a reinforcement learning agent, LastOrder,  
 that can select both macro and micro actions with high accuracy. They employ a LSTM based network to deal with partial observability while using Ape-X DQG double Q-learning to handle reward sparsity.  LastOrder achieves 83\% win rate vs.\ the AIIDE 2017 StarCraft AI Competition bot set, and placed 11th out of 25 in the 2018 competition.

To date the most successful system built on SC2 is AlphaStar~\cite{alpha} which achieved grandmaster level, beating expert human players in all three races (Protoss, Zerg, and Terran) online; though it has been defeated by a human, it is ranked over 99.8\% of human players. AlphaStar uses a complex combination of machine learning methods; after being initialized using supervised learning from replays of human players, it trains using a population-based reinforcement learning process in which basic agents compete against each other in simulated StarCraft tournament ladder.
Basic macro and micro strategies are acquired from supervised learning on replays, and new strategies emerge during the tournament process which is carefully managed to avoid extincting valuable strategies. 
The underlying neural representation includes both a transformer and deep LSTM core, but it is difficult to draw conclusions about their relative contributions. 

\subsection{Build Order Prediction}
Deep learning methods have also been successfully applied to supervised prediction tasks, such as build order prediction. Doing this requires a large replay dataset harvested from online repositories~\cite{msc,stardata}; ideally it should contain data from players with varying ranges of expertise. Our research builds on the Macromanagement for SC2 (MSC) dataset created by Wu et al.\cite{msc} which is the largest one available. 

\begin{figure} [htb]
    \centering
    \includegraphics[width=0.8\columnwidth]{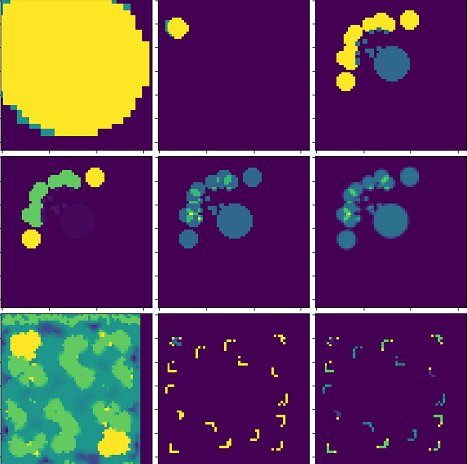}
    \caption{Visualization of spatial features extracted from the MSC dataset:
    1) map height, 2) player relative, 3) unit type, 4) unit density, 5) minimap player relative, and 6) screen creep.}
    \label{fig:sc2}
\end{figure}

MSC consists of 36,619 high quality replays, extracted from 64,369 games after pre-processing and parsing to ensure data quality. Moreover, it contains the final results of each game at every time step, making it suitable for win/loss prediction. Feature vectors (101 elements) contain state-action pairs in normalized form, along with the game result; this representation makes it a better choice for evaluating global state prediction than StarData which only contains a small number of feature vectors annotated with the final game result. A key benefit of this dataset is the inclusion of spatial features (Figure~\ref{fig:sc2}). The dataset comes with predefined training, validation and testing sets.

Build order prediction can be modeled as classification problem rather than a sequence to sequence learning one, as is done in the paper  \cite{risi} which use a 5-layer fully connected network with softmax probability output to predict the build order from a given game state. The input to their network is a 210-element vector including normalized features such as available resources, buildings under construction, opponents' buildings, and the available supplies. Their architecture has a higher top-1 prediction error of 54.6\% which is unsurprising given that it is unable to leverage sequential information.  
We benchmark our research against the best performing method proposed by \cite{msc}. Their basic network consists of fully connected layers, followed by Gated Recurrent Units (GRU), followed by another fully connected layer and a sigmoid output for predicting the final outcome of the game.
The network is optimized using Adam optimizer and binary cross entropy loss function. This model predicts the global state with a best accuracy of 61.1\% for Terran vs. Terran games. They also introduce a two-branch version of the network.   The upper branch extracts spatial features from mini-maps while the lower branch extracts information from global features. These features are then combined and fed into the linear layer followed by a GRU and another linear layer with a softmax probability output. This version of the network achieves the best build order accuracy of 74.9\% for Zerg vs. Zerg games while achieving global state prediction of 57.8\% for Protoss vs Protoss games.
Like \cite{msc}, we employ both the global and spatial features as input using a two branch network; however we introduce the use of a transformer as a superior replacement for the GRU for capturing temporal dependencies.  Also our proposed architecture simultaneously performs both prediction tasks using a multi-headed output. 


\section{Data}

\begin{figure*}[ht]
    \centering
    \includegraphics[width=.8\textwidth]{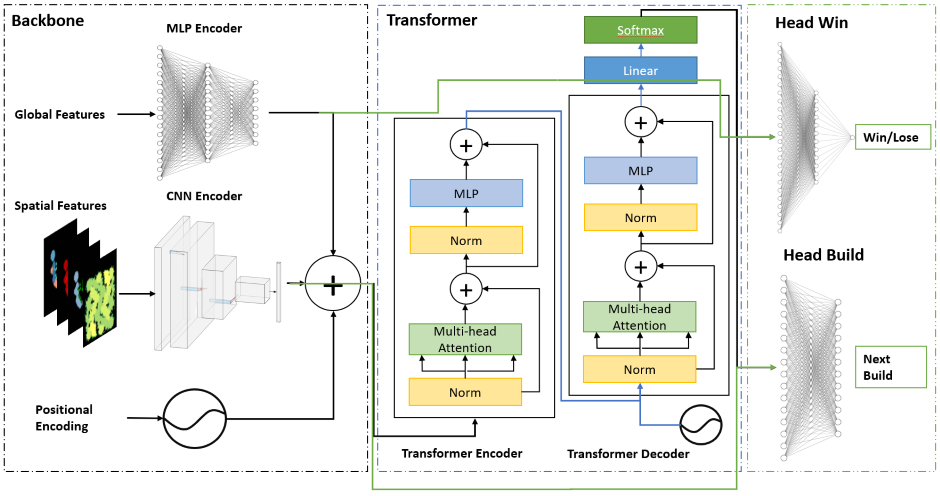}
    \caption{Full Model Overview. Global features are fed to an MLP encoder and spatial features to a CNN encoder. The output from both encoders is aggregated with a positional encoding and sent to a transformer encoder and decoder. Our architecture contains six layers of transformer
encoders and decoders (only one is shown). Unlike a vanilla transformer, the transformer decoder is also provided with the positional encoding. A skip connection from the MLP encoder is added to the output of the transformer stack which is fed to Head Win. Similarly, a skip connection is used to connect the CNN encoder to Head Build. The ablated version of the model lacks skip connections and decoder layers.}
    \label{fig: Model Overview }
\end{figure*}

The MSC dataset is based on SC2LE \cite{chal} which uses the SC2 Linux Packages 3.16.1. The dataset contains over 36,000 preprocessed high quality SC2 game replays. Table I shows the number of replays of each game type. These replays contain games played by skilled SC2 players. The dataset is divided into training, validation, and test sets in the ratio 7:1:2, facilitating comparison betwen methods. Each replay is guaranteed to have:
\begin{compactitem}
    \item At least 10,000 frames
    \item Over 10 actions per minute (APM) for both players
    \item A matchmaking ratio (MMR) of greater than 1000 (low matchmaking ratios may occur due to broken replays or weak players).
\end{compactitem} 

\begin{table} [h!]
\label{table:1}
\caption{Number of MSC replays for each racial matchup (post-processing).  Note that some of the matchups have a significantly larger number of training examples than others which has implications on transfer learning performance.}
\centering
\renewcommand{\arraystretch}{1.5}
\begin{tabularx}{0.45\textwidth} { 
  >{\raggedright\arraybackslash}X 
  | >{\centering\arraybackslash}X}
 \textbf{Matchup} & \textbf{No. of Replays} \\
 \hline
  Terran vs Terran & 4897 \\ 
  \hline
  Terran vs Protoss & 7894\\ 
  \hline
  Terran vs Zerg & 9996\\
  \hline
  Protoss vs Protoss & 4334\\
  \hline
  Protoss vs Zerg & 6509\\
  \hline
  Zerg vs Zerg & 2989\\
\end{tabularx}

\end{table}

Each replay contains global features, spatial features, cumulative scores, win/loss outcome, and ground truth actions~\cite{msc}.  The global feature vector consists of the following sub-vectors: 1) frame id, 2) user's resource collection, 3) alerts received by the player, 4) upgrades applied by the player, 5) techniques researched by the player, 6) units and buildings owned by the player, and 7) enemy buildings and units observed by the player. These features are normalized in the interval [0, 1]. The spatial feature vector consists of the following features: 1) screen features, 2) mini-map features, 3) map height and width, 4) unit type, 5) unit density, and 6) unit relative. Fig.~\ref{fig:sc2} depicts some of the spatial features. The dimension of these features is $\mathbb{R}^{13\times64\times64}$. Similar to global features, these features are also normalized to [0, 1]. Additional features include: 1) game reward i.e., final result (0 for loss and 1 for win), 2) cumulative score (not normalized), and 3) ground truth actions. Table~\ref{table:2} provides details on the action options and unit types available for each race.
\begin{table}
\caption{Action options by race.}
\label{table:2}
\centering
\renewcommand{\arraystretch}{1.5}
\begin{tabularx}{0.45\textwidth} { 
  >{\raggedright\arraybackslash}X 
  | >{\centering\arraybackslash}X 
  | >{\centering\arraybackslash}X }
 \textbf{Race Type} & \textbf{\# Actions} & \textbf{Unit Types}\\ 
 \hline
 Terran & 75 & 336 \\
 \hline
 Protoss & 61 & 246 \\
 \hline
 Zerg & 74 & 714 \\
\end{tabularx}

\end{table}

\section{Method}
Figure~\ref{fig: Model Overview } shows our proposed neural architecture. The backbone of the model consists of two branches - one for extracting information from global features and one for creating a compact representation of spatial features.  A multi-layer perceptron (MLP) encoder, consisting of 3 linear layers with rectified linear units (ReLU), is used to learn a representation of the global features with a dimension of $\mathbb{R}^{256}$. Similarly, a convolutional neural network (CNN) encoder with 3 Conv2D layers followed a max pooling layer and ReLU activation is used to learn a compact representation of the spatial features.  These representations of game states are then fed to a transformer encoder together with the positional encoding that preserves the ordering information. Following the discussion of \cite{attn_aug} and \cite{img_trans}, we adopt the fixed positional encodings since the transformer model is permutation invariant. 

The architecture of the transformer encoder is very similar to the standard encoder proposed by \cite{attn} where each layer consists of a layer normalization module, a multi-headed self-attention module, and a feed forward network (FFN) module along with residual connections. The transformer encoder ingests a sequence of elements consisting of: 1) the output from the MLP encoder, 2) the output from CNN encoder (after being flattened to match the dimension of output of MLP encoder) and 3) positional encoding.  

 The transformer decoder also follows the standard implementation; however, we also include the positional encoding along with the sequence produced by the transformer encoder. The decoder receives the aggregated input consisting of 1) transformer encoder output, and 2) positional encoding.  The softmax output from transformer decoder is fed to two separate prediction heads: Head Win and Head Build.

Our architecture contains six layers of each transformer encoder and transformer decoder with $d_{model}$ set to $\mathbb{R}^{256}$, number of self-attention heads to eight, a feed-forward dimensionality of $\mathbb{R}^{1024}$, dropout of 0.5 and ReLU activation. 

\par Head Win is a 2-layer FFN with ReLU non-linearity, using a sigmoid output. It predicts the final result of the game (win or loss) given the current resources. This output is used to measure the accuracy of Global State Prediction (GSP).  Like Head Win, Head Build is a 2-layer FFN with ReLU non-linearity and softmax output; it is used to evaluate the accuracy of Build Order Prediction (BOP).

Our full model also utilizes skip connections. The idea behind skip connections is to feed the output of a layer directly to a deeper layer while skipping a few layers in between. Skip connections provide some known advantages: 1) creation of an alternative gradient path, 2) prevention of the gradient vanishing problem, and 3) assistance with model convergence. Previous work has shown the effectiveness of skip connections in deep computer vision networks~\cite{resnet,unet}.  In our model, one skip connection is taken from the MLP encoder and is added to the output of the transformer before feeding to Head Win. The second skip connection is taken from the output of the CNN encoder and added to the output of the transformer to be fed to Head Build.  

We compare our architecture to the highest performing baseline described by \cite{msc} which uses a model that leverages both global and spatial features. Their model consists of couple of Conv2D layers to extract spatial features, a linear layer to extract global features; after combination, features are passed to a gated recurrent unit (GRU) with a hidden dimension of $\mathbb{R}^{128}$. A sigmoid output is used for global state prediction, and a softmax one for build order prediction.


\section{Experiments}

We implement our model in the PyTorch library.  The model is trained with Adam optimizer~\cite{adam} by setting initial learning rate to $1 \times 10^{-3}$, $\beta_1 = 0.9$ and $\beta_2 = 0.999$. The learning rate is reduced by half every second epoch.  The model was trained for 10 epochs with a batch size of 20 replays over 10 time steps on a Nvidia Tesla V100 GPU for each game. 
For Head Win, we select the same loss function as was employed by the GRU baseline i.e., the Binary Cross Entropy:
\begin{equation}
\label{eq:1}
\begin{split}
    \mathcal{L}_{global} = -\log(P(R = 1 | \Omega_t)) R_t \\ - \log(P(R = 0 | \Omega_t))(1 - R_t)
\end{split}
\end{equation}
where $\Omega_t$ represents the observations at time $t$ and $R$ represents the final result of the game. 

Unlike the baseline which employs Negative Log Likelihood Loss (NLL), Head Build employs the Cross Entropy Loss function. In this case, the loss function is given by the following equation:
\begin{equation}
\label{eq:2}
\begin{aligned}
    \mathcal{L}_{build} = -\sum_{i=1}^{N} a_i \log(P(a_i | \Omega_t))  
\end{aligned}
\end{equation}
where $a_i$ is the ground truth action while $P(a_i | \Omega_t)$ represents the predicted action.

The final loss for our model is given by the combination across both tasks:

\begin{equation}
\label{eq:3}
\begin{aligned}
    \mathcal{L} = \mathcal{L}_{global} + \mathcal{L}_{build}
\end{aligned}
\end{equation}
In contrast, the baseline is done as two separate models with individual loss functions.


\section{Results}
This section presents a comprehensive evaluation of our proposed transformer vs.\ the best performing version of the baseline on two macromanagement prediction tasks.  We also evaluate ablated versions of our model to show the relative contributions of the different components.   Finally we illustrate the generalization power of the model in a transfer learning setting.

Both Global State Prediction (\textbf{GSP}) and Build Order Prediction (\textbf{BOP}) are challenging aspects of SC2 macromanagement. GSP looks at the game's current state and tries to predict the outcome of the game (win or loss)~\cite{gsp}. BOP predicts the next production action (building, training, or researching) based on current state~\cite{bop}.  Interestingly, all models perform better at BOP vs. GSP, despite the fact that a random chance guesser would perform better at predicting win/loss vs. predicting the next build action.  We evaluate mean accuracy on both those tasks across all time steps using the provided test/train split in MSC.

Note that \cite{msc} actually trained separate models for the GSP and BOP tasks; in contrast we train a single model with multi-headed output.
Our proposed transformer model captures the temporal relationships and context necessary to achieve better prediction power, and thus outperforms the baseline GRU model~\cite{msc} on both tasks. Table~\ref{table:3} provides a detailed breakdown of our results  for every racial matchup. 

\begingroup

\renewcommand{\arraystretch}{1.6}
\begin{table}[h]
\caption{Mean accuracy for global state and build order prediction. Our proposed transformer-based model achieves improvements on both tasks over the GRU baseline across the total dataset and in most matchups. }
\label{table:3}
\centering

\begin{tabularx}{0.45\textwidth} { 
  >{\raggedright\arraybackslash}X 
  | >{\centering\arraybackslash}X 
  | >{\centering\arraybackslash}X 
  | >{\centering\arraybackslash}X 
  | >{\centering\arraybackslash}X }
 
 \multirow{2}{4em}{\textbf{Games}} & \multicolumn{2}{c |}{\textbf{Baseline (GRU)}} & \multicolumn{2}{c}{\textbf{Our (Transformer)}}\\  
 \cline{2-5}
 & \textbf{GSP} & \textbf{BOP} & \textbf{GSP} & \textbf{BOP} \\
 \hline
 TvT & 50.9 & 73.1 & \textbf{52.23} & \textbf{74.38} \\
 \hline
 PvP & 57.8 & 74.2 & \textbf{58.42} & \textbf{74.6} \\
 \hline
 ZvZ & 54.7 & \textbf{74.9} & \textbf{56.01} & 74.6 \\
 \hline
 PvT & 57.0 & 69.6 & \textbf{58.63} & \textbf{77.58} \\
 \hline
 PvZ & \textbf{56.9} & 74.2 & 56.19 & \textbf{77.92} \\
 \hline
 TvZ & \textbf{56.1} & 74.8 & 55.79 & \textbf{75.22} \\
\end{tabularx}

\end{table}
\endgroup


\subsection{Skip Connection}
To evaluate the contributions of different aspects of our proposed neural architecture, we conduct an ablation study on the performance of different design choices. Our full model contains skip connections which bypass the transformer encoder and decoder stacks, providing the Head Win and Head Build with direct access to the original state vectors. 
Table~\ref{table:4} shows the results of removing those skip connections. Skip connections do provide a slight advantage; our full model gained a couple of points over the one without skip connections on both tasks.  

\begingroup
\renewcommand{\arraystretch}{1.6}
\begin{table}[h!]
\caption{Transfer learning often further improves the performance of our transformer model, compared to training from scratch.}
\label{table:5}
\centering
\begin{tabularx}{0.45\textwidth} { 
  >{\centering\arraybackslash}p{0.07\textwidth} 
  | >{\centering\arraybackslash}p{0.1\textwidth} 
  | >{\centering\arraybackslash}p{0.08\textwidth}
  | >{\centering\arraybackslash}p{0.1\textwidth}}
 
 \multicolumn{2}{c|}{\textbf{Our Model}} & \multicolumn{2}{c}{\textbf{With Transfer Learning}} \\
 \hline
  \multirow{2}{4em}{PvP} & \textbf{GSP: 58.42} & 
  \multirow{2}{4em}{TvT to PvP} & GSP: 56.01 \\
 & BOP: 74.6 & & \textbf{BOP: 76.57} \\
 \hline
 \multirow{2}{4em}{TvT} & GSP: 52.23 &
 \multirow{2}{4em}{PvP to TvT} & \textbf{GSP: 55.61} \\
 & BOP: 74.38 & & \textbf{BOP: 75.11} \\
 \hline
 \multirow{2}{4em}{PvT} & \textbf{GSP: 58.63} &
 \multirow{2}{4em}{TvT to PvT} & GSP: 51.64 \\
 & BOP: 77.58 & & \textbf{BOP: 78.19} \\
 \hline
 \multirow{2}{4em}{TvT} & GSP: 52.23 &
 \multirow{2}{4em}{PvT to TvT} & \textbf{GSP: 55.54} \\
 & BOP: 74.38 & & \textbf{BOP: 74.59} \\
 \hline
 \multirow{2}{4em}{ZvZ} & GSP: 56.01 &
 \multirow{2}{4em}{PvP to ZvZ} & \textbf{GSP: 58.3} \\
 & BOP: 74.6 & & \textbf{BOP: 75.98} \\
 \hline
 \multirow{2}{4em}{PvZ} & \textbf{GSP: 56.19} &
 \multirow{2}{4em}{PvZ to PvP} & GSP: 54.47 \\
 & \textbf{BOP: 77.92} & & BOP: 76.49 \\

\end{tabularx}

\end{table}
\endgroup

\subsection{Transformer Decoder}
One research question that concerned us was whether the transformer decoder stack was redundant, since our proposed architecture includes output heads for computing win/loss and next build.  Unlike many translation architectures that employ transformers, the loss function is not applied directly to the output of the decoder. Table~\ref{table:4} shows the results of omitting the decoder stack, which significantly reduces the number of model parameters. We observe that including the decoder stack results is small improvements in both prediction tasks.  

\subsection{Self-Attention Model}
One drawback of the transformer is that it requires a significant computational investment to train the large number of parameters. We hypothesized that a simpler self-attention-based model may be able to capture contextual dependencies in SC2 production; for instance actions in build trees commonly co-occur with other actions. Similar to our transformer-based model, the self-attention model also consists of two branches: 1) an MLP encoder, and 2) a CNN encoder. We added a separate self-attention module for each branch. The output from both self-attention modules is then aggregated and is fed to 2-layer FFN which has two output heads: 1) one for global state and 2) one for build order prediction.  

\begingroup
\renewcommand{\arraystretch}{1.6}
\begin{table*}[!h]
\caption{Our full model achieves improvements on both tasks compared to the ablated versions of our model.}
\label{table:4}
\centering
\begin{tabularx}{0.8\textwidth} { 
   >{\centering\arraybackslash}X |
  >{\centering\arraybackslash }X 
  | >{\centering\arraybackslash }X 
  | >{\centering\arraybackslash }X
  | >{\centering\arraybackslash }X
  | >{\centering\arraybackslash }X 
  | >{\centering\arraybackslash }X 
  | >{\centering\arraybackslash }X
  | >{\centering\arraybackslash }X}
 
  \multirow{2}{4em }{\textbf{Games}} & 
 
  \multicolumn{2}{c |}{\textbf{Full Model}} & \multicolumn{2}{c |}{\textbf{No Skip Connection}} & 
  \multicolumn{2}{c |}{\textbf{No Decoder}} & 
  \multicolumn{2}{c}{\textbf{Self-Attention Only}}\\  
\cline{2-9}
 & \textbf{GSP} & \textbf{BOP} & \textbf{GSP} & \textbf{BOP} & 
 \textbf{GSP} & \textbf{BOP} & \textbf{GSP} & \textbf{BOP}\\
 \hline
Averaged  &
\textbf{55.43} & \textbf{75.98} & 52.04 & 74.06 & 53.4 & 74.97 & 23.21 & 24.97
\end{tabularx}
\end{table*}
\endgroup

Table~\ref{table:4} shows that this version of the model performed surprisingly badly, underperforming both the transformer and the GRU baseline.
This indicates that in some ways the recency bias of the GRU is very useful since it focuses the model on a valuable temporal segment.  In contrast the self-attention model fails to use temporal information effectively (disproving \textbf{H2}). The basic self-attention 
 module also lacks the positional encoding component of the transformer which helps the model learn ordering dependencies.  

\subsection{Transfer Learning} 
One benefit of the transformer model is its strong ability to generalize to new problems.  We demonstrate this by training our proposed model in a transfer learning setting in which weights are learned from one racial matchup and transferred to another.  To do this, we take a trained model from one pairing, freeze the MLP encoder, the CNN encoder, and the transformer, reinitialize both heads, train this model for 3 to 5 epochs and record the accuracy for both global state and build order. We keep the other training parameters fixed except the learning rate which is reduced to $1 \times 10^{-5}$ before starting the transfer learning experiments. 
\par With transfer learning, our approach achieves even better performance in less time than the model trained from scratch on the same game. Table~\ref{table:5} presents the findings of our transfer learning experiments. In cases where both datasets share a common race, our model achieves even better results.

\section{Conclusion and Future Work}
This paper introduces a new neural architecture for macromanagement prediction tasks in SC2; our transformer-based architecture outperforms the GRU baselines on mean accuracy of global state and build order prediction on the MSC dataset.  We demonstrate that the transformer is a superior option for capturing long-term temporal dependencies compared to recurrent neural networks which suffer from a recency bias (\textbf{H1}).  However, the GRU is decisively superior to a simple self-attention system, disproving \textbf{H2}.  Our ablative study justifies the inclusion of skip connections and transformer decoders. Our transformer is similar to the original one but includes a positional encoding input for the decoders.  Our results on a transfer learning problem indicate that the model  generalizes well (\textbf{H3}).
One of the strengths of the transformer is its ability to parallelize learning, unlike recurrent neural networks. Unfortunately the MSC dataset is not well designed for parallel loading, and we plan to tackle this problem in future work through the creation of a new dataset.
Another promising avenue for future research is coupling our transformer architecture with an RL macromanagement agent.  

\section{Acknowledgments}
This research was partially supported by ARL STRONG W911NF-19-S-0001.


\bibliographystyle{IEEEtran}
\bibliography{main.bib}
\end{document}